\pdfoutput=1

\documentclass[11pt]{article}

\usepackage[final]{acl}
\usepackage{booktabs}
\usepackage{threeparttable}
\usepackage{times}
\usepackage{latexsym}
\usepackage{svg}

\usepackage{todonotes}
\makeatletter
\newcommand*\iftodonotes{\if@todonotes@disabled\expandafter\@secondoftwo\else\expandafter\@firstoftwo\fi} 
\makeatother

\newcommand{\af}[1]{\textit{#1}}

\usepackage[T1]{fontenc}

\usepackage[utf8]{inputenc}

\usepackage{microtype}

\usepackage{inconsolata}

\usepackage{graphicx}

\newenvironment{enumerate*}%
  {\begin{enumerate}%
    \setlength{\itemsep}{0.9pt}%
    \setlength{\parskip}{0.9pt}%
    \setlength{\topsep}{0.9pt}}%
  {\end{enumerate}}

\usepackage{xcolor}         
\usepackage{colortbl}         
\usepackage{amssymb} 
\usepackage{enumitem}
\usepackage{booktabs} 

\usepackage[most]{tcolorbox}
\tcbuselibrary{listingsutf8}
\usepackage{listings}
\usepackage{xcolor}

\definecolor{darkgray}{rgb}{0.225, 0.225, 0.225}
\definecolor{varcolor}{rgb}{0.4, 1.0, 1.0}    %
\definecolor{bracecolor}{rgb}{0.6, 0.8, 1.0}  %

\lstdefinestyle{mypython}{
    backgroundcolor=\color{darkgray},
    basicstyle=\ttfamily\footnotesize\color{white},
    keepspaces=true,
    breaklines=true,
    breakatwhitespace=false,
    breakindent=0pt,
    breakautoindent=false,
    prebreak=\mbox{},
    postbreak=\mbox{},
    columns=fullflexible,     %
    showstringspaces=false,
    frame=none,
    language={},
    keywordstyle=,
    stringstyle=,
    commentstyle=,
    moredelim=**[is][\color{varcolor}]{@@}{@@},
    moredelim=[s][\color{bracecolor}]{\{\{}{\}\}},
    literate={{@}{}{1}},
}

\newtcblisting{promptbox1}{
    title=\texttt{Evidence Extractor},
    listing only,
    listing options={style=mypython},
    colback=darkgray,
    colframe=gray!60,
    colbacktitle=gray!30,
    coltitle=black,
    fonttitle=\bfseries,
    enhanced,
    frame hidden=false,
    boxrule=0.5pt,
    arc=1mm,
    left=2mm,
    right=2mm,
    top=1mm,
    bottom=1mm
}

\newtcblisting{promptbox2}{
    title=\texttt{Index Navigator},
    listing only,
    listing options={style=mypython},
    colback=darkgray,
    colframe=gray!60,
    colbacktitle=gray!30,
    coltitle=black,
    fonttitle=\bfseries,
    enhanced,
    frame hidden=false,
    boxrule=0.5pt,
    arc=1mm,
    left=2mm,
    right=2mm,
    top=1mm,
    bottom=1mm
}

\newtcblisting{promptbox3}{
    title=\texttt{Tabular Validator},
    listing only,
    listing options={style=mypython},
    colback=darkgray,
    colframe=gray!60,
    colbacktitle=gray!30,
    coltitle=black,
    fonttitle=\bfseries,
    enhanced,
    frame hidden=false,
    boxrule=0.5pt,
    arc=1mm,
    left=2mm,
    right=2mm,
    top=1mm,
    bottom=1mm
}

\newtcblisting{promptbox4}{
    title=\texttt{Code Reconciler},
    listing only,
    listing options={style=mypython},
    colback=darkgray,
    colframe=gray!60,
    colbacktitle=gray!30,
    coltitle=black,
    fonttitle=\bfseries,
    enhanced,
    frame hidden=false,
    boxrule=0.5pt,
    arc=1mm,
    left=2mm,
    right=2mm,
    top=1mm,
    bottom=1mm
}

\title{\textit{Code Like Humans}: A Multi-Agent Solution for Medical Coding}

\author{
  \textbf{Andreas Motzfeldt\textsuperscript{1,2}}\quad\quad
  \textbf{Joakim Edin\textsuperscript{1,3}}\quad\quad
  \textbf{Casper L. Christensen\textsuperscript{1}}\\[0.15cm]
  \textbf{Christian Hardmeier\textsuperscript{2}}\quad\quad
  \textbf{Lars Maaløe\textsuperscript{1}}\quad\quad
  \textbf{Anna Rogers\textsuperscript{2}}\\[0.3cm]
  \textsuperscript{1}Corti.ai\quad
  \textsuperscript{2}IT University of Denmark\quad
  \textsuperscript{3}University of Copenhagen
}

\begin{document}
\maketitle

\begin{abstract}
In medical coding, experts map unstructured clinical notes to alphanumeric codes for diagnoses and procedures. We introduce \textit{Code Like Humans}: a new agentic framework for medical coding with large language models. It implements official coding guidelines for human experts, and it is the first solution that can support the full ICD-10 coding system (+70K labels). It achieves the best performance to date on rare diagnosis codes (fine-tuned discriminative classifiers retain an advantage for high-frequency codes, to which they are limited). Towards future work, we also contribute an analysis of system performance and identify its `blind spots' (codes that are systematically undercoded). 
\end{abstract}

\vspace{0.5em}
{\footnotesize \noindent\textbf{Code:} \href{https://github.com/MotzWanted/codeseeker}{https://github.com/MotzWanted/codeseeker}}

\section{Introduction}

For statistical and billing purposes, unstructured clinical notes need to be mapped to medical codes: alphanumeric codes of diagnoses or procedures~\cite{chandawarkar_revenue_2024}. The International Classification of Diseases (ICD) is the most widely used system for diagnosis codes. Unfortunately, it is a time-intensive task,\footnote{It takes a professional inpatient human coder approximately 30 minutes per case on average~\cite{chen_automatic_2021}.} in which errors can cause patient mistreatment and lost revenue~\cite{gao_optimising_2024, gaffney_medical_2022}. %

\begin{figure}[htbp]
  \centering
  \includeinkscape[width=\linewidth]{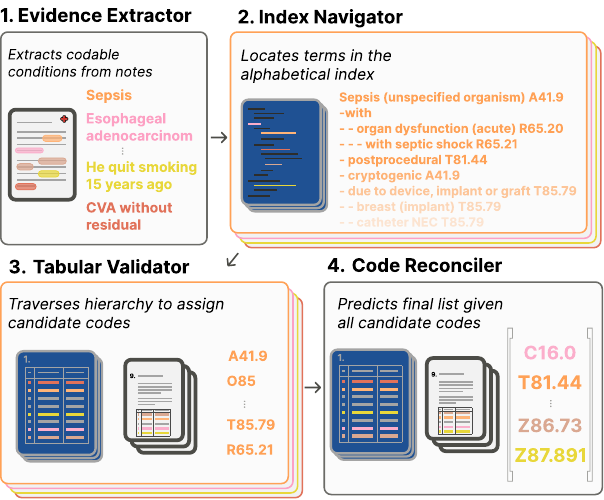}
  \caption{Overview of \textit{Code-Like-Humans}, our agentic framework whose structure mirrors the \textit{Analyze-Locate-Assign-Verify} approach of the UK National Health Service. The four agents sequentially emulate how medical coders extract evidence, navigate the alphabetical index, validate the ICD hierarchy, and reconcile coding conventions to `translate' clinical notes into ICD codes.}
\label{fig:agent_system}
\end{figure}

Progress in NLP methods for medical coding has been stagnant in recent years. There have been several attempts to apply large language models (LLMs)~\cite{boyle_automated_2023, falis_can_2024}, but they have yet to yield improvements over the 2022 state-of-the-art~\cite{huang_plm-icd_2022} that used the BERT architecture~\cite{devlin_bert_2019}.

We hypothesize that one reason for the lack of progress may be the disconnect between current modeling approaches and how human medical coders work~\cite{gan_aligning_2025}. Human coders begin not from memory but from the \emph{alphabetical index} of ICD, a list that exhaustively maps clinical terms to codes capturing synonyms, eponyms, and contextual cues. Although this resource is an authoritative source of information used by medical coders to narrow down the search space, the current approaches do not consider it in any way.

To address this, we propose \textit{CLH} (Code Like Humans): an LLM-based agentic framework\footnote{An LLM-based agentic system expands the functionality of a stand-alone LLM in a goal-oriented manner to plan and take actions over time, interacting with tools, data sources, or other agents~\cite{qiu_llm-based_2024}.} that leverages ICD resources such as the alphabetic index. \textit{CLH} is inspired by the medical coding workflow developed for healthcare professionals in the UK~\cite{nhs_england_national_2023, cms_icd-10-cm_2025}. Our approach is summarized in \autoref{fig:agent_system}.

Our contributions are:
\begin{enumerate}
\item We propose \textit{CLH}, an LLM-based agentic framework designed to mirror human medical coders by leveraging the same external resources: the alphabetic index, the ICD hierarchy, and the ICD guidelines.
\item We develop and publicly release an agentic implementation of \textit{CLH} that can support the complete US ICD-10 system (70K labels). This solution achieves comparable performance with the state-of-the-art fine-tuned classifiers on rare codes, although the frequent codes still pose a challenge. 
\item We provide an extensive analysis of our solution, identifying its difficulties in several areas of medical taxonomy for future work. %
\end{enumerate}

We conclude by discussing the overall readiness of the current solutions for medical coding. While in NLP this task has historically been approached from the \textit{automation} perspective (i.e. end-to-end classification), at this stage, none of the existing approaches are ready for real-world deployment as a replacement for human coders. We argue that the task should generally be reconsidered as \textit{assisting} the human coders, based on the identification of the pain points in their process that LLM technology can realistically help with. %

\section{Related Work and Novelty}
\label{sec:related}
Current approaches rely on encoder-decoder architectures that embed clinical notes and codes into separate embedding spaces. Label-wise attention with fine-tuning is typically used to align them~\cite{mullenbach_explainable_2018, li_icd_2019}. Prior work has sought to improve alignment by initializing embeddings with textual descriptions \cite{dong_explainable_2021}, synonyms \cite{yuan_code_2022}, or code co-occurrence \cite{xie_ehr_2019, cao_hypercore_2020}. However, state-of-the-art results have been achieved using discriminative models, pre-trained on biomedical texts and then fine-tuned for classification (PLM-ICD)~\cite{huang_plm-icd_2022, edin_automated_2023}. This is the main baseline considered in this work.

These methods face three main challenges, listed below, together with our proposed solutions. 

\paragraph{(i) Extremely large label set.} The full ICD-10 coding system has over 70K possible labels to assign. In most of the prior work \cite[e.g.][]{mullenbach_explainable_2018, li_icd_2019, vu_label_2020, cao_hypercore_2020, yuan_code_2022}, label-wise attention is filtered to codes present in MIMIC, ~10\% of the ~70K ICD-10-CM codes. There are currently no public datasets that could verify the effectiveness of any approach for the full ICD-10 space. Furthermore, many prior studies report exclusively the performance on the top 50 codes~\cite{gan_aligning_2025}. This setup simplifies the actual problem of medical coding, but it does not accurately reflect the needs of a real-world application.  

\textit{Our solution.} \textit{CLH} retrieves codes from the ICD alphabetic index, enabling true open-set coding at inference. This makes \textit{CLH} the first approach to cover\footnote{\textit{CLH} can, in theory, assign any of the ~70K codes. However, this does not imply uniform coverage in practice, since a specific implementation may have `blind spots' that should be established empirically via performance analysis, as in other classification problems, especially under class-imbalanced.} the full ICD-10 solution space.  

\paragraph{(ii) Predicting codes with few or no training examples.}
When discriminative models are fine-tuned with standard cross-entropy on imbalanced data, random mini-batches mirror the label frequencies in the training set. The learned decision boundaries are biased toward frequent codes, under-predicting the rare codes. This bias might also lead to poor performance when code distributions differ across clinical settings (unfortunately, there are no public datasets that could be used to estimate the scope of this problem). Thus, such models only predict codes observed during training. Extending label-wise attention to the full solution space would result in negligible scores for unseen codes due to untrained weights, which is why in practice, the label set is a subset (see (i)). Recent works explored LLMs without fine-tuning for code extraction~\cite{yang_multi-label_2022, boyle_automated_2023, gero_self-verification_2023}, data augmentation for rare codes~\cite{falis_can_2024}, and few-shot generative coding with gains on a constructed few-shot split of MIMIC-III~\citep{yang_multi-label_2022}, but none surpassed PLM-ICD.

\textit{Our solution.} Rather than learning a label-frequency prior from training data, \textit{CLH} uses a retrieval-induced distributional prior from the Alphabetical index, which enables probability mass on codes with few or zero training examples.

\paragraph{(iii) Processing long inputs.} BERT-style models necessitate chunking notes due to limited input length (512 tokens), which complicates optimization~\cite{pascual_towards_2021}. Chunking converts document-level supervision into a multi-instance setting where aggregators such as max-pooling pass gradients only through the highest-scoring segment and cross-segment dependencies are easily missed. Medical notes can be fairly long, up to 8,500 tokens in MIMIC dataset \cite{johnson_mimic-iii_2016, johnson_mimic-iv_2023}. Although LLMs handle longer inputs better, very long contexts still pose challenges~\cite{karpinska_one_2024, li_long-context_2024,kim2025rulermeasureallbenchmarking}. Even for most LLMs,\footnote{Gemini models handle up to 1M tokens~\cite{lee_can_2024}, yet to date most evaluations use the needle-in-the-haystack paradigm~\cite{kamradt2023needle}, e.g. ~\cite{hsieh2024ruler}, where they perform much better than on a task requiring reasoning over an entire book~\cite{karpinska_one_2024}.} providing the entire ICD solution space as in-context retrieval~\cite{lee_can_2024} is not computationally feasible, as it easily would exceed 1M tokens. It is also not realistic: human coders narrow the solution space using multi-step processes guided by the alphabetical index and official guidelines~\cite{dong_automated_2022, gan_aligning_2025}.

\textit{Our solution.} \textit{CLH} adopts an agentic approach, which decomposes the processing of long context as a sequential task, where the system searches, verifies, and predicts one code at a time based on guidelines. This avoids having to process extremely long context in one go.

\section{Background: a Gentle Introduction to Medical Coding}
\label{sec:methodology}
Human medical coders are healthcare professionals with extensive training in one or more medical coding systems.\footnote{According to \citet{otero_varela_international_2024}, the training for a medical coding certification takes several months.}. Given some clinical documentation (progress notes, discharge summaries, etc.), they must choose among the thousands of codes the most specific one that applies to this case. We will use the example of the US modification of ICD-10 (ICD-10-CM)~\cite{cms_icd-10-cm_2025}, which comprises over 70,000 codes. Example snippets from the ICD hierarchy and alphabetical index can be seen in the \autoref{app:external_modules}.
    
These codes are defined in a tree-structured hierarchy where the proximity of the codes indicates some similarity.\footnote{In ICD, this hierarchy is referred to as the `tabular list'.} The alphanumeric characters in a medical code refer to its position in the hierarchy: e.g., in the code \textit{A22.7}, ``A'' points to a chapter, ``22'' points to a category, and the numbers after the punctuation indicate the conditions with increasing level of specificity. Only the most specific possible code can be assigned. However, proximity in the hierarchy does not guarantee clinical similarity. For example, `sepsis' is mentioned 39 times in multiple chapters, and it corresponds to codes 
\textit{A22.7} 
and \textit{T81.44}, among many others. %
Therefore, finding the correct code by traversing from the top of the hierarchy to the bottom is difficult.

To resolve such cases, the coders must also use the alphabetical index, which provides information about the context in which a given code is appropriate. For example, it could help the coder to decide between \textit{A22.7} and \textit{T81.44} codes for `sepsis', because the former is listed under ``anthrax'' context, and the latter under ``postprocedural''.

The alphabetical index alone is also insufficient because it does not necessarily point to the most specific code (i.e., the leaf node in the ICD tree). Hence, the coder must verify that the candidate code is the most specific code possible for this case in the ICD hierarchy.

The output of medical coding is not a single code, but a list of applicable codes. The order of those codes may itself be meaningful in some cases. There are also rules for precedence between specific codes, e.g., the code for ``Alzheimer's'' should come before the code for ``dementia''.

All human coders must be familiar with official ICD guidelines to ensure accurate and comprehensive coding. The current ICD-10-CM guidelines are 115 pages long. These guidelines serve as the manual for medical coding and provide instructions on using the alphabetical index and navigating the hierarchy. Additionally, they provide chapter-specific rules on selecting and combining codes, which are fundamental to clarifying otherwise ambiguous choices among several candidate codes.

\section{Code Like Humans (\textit{CLH}) Framework}
\subsection{Architecture}
\label{sec:architecture}
The proposed \textit{CLH} framework decomposes the medical coding task into four steps, corresponding to the \textit{Analyze-Locate-Assign-Verify} approach implemented by the UK National Health Service~\cite{nhs_england_national_2023}. We first describe the framework abstractly in terms of the interface and functionality of each component. In section \ref{sec:implementation}, we present our implementation, which reflects one of many possible realizations of \textit{CLH} and is not intended as a definitive standard. 

\paragraph{Step 1: \af{evidence extractor}.} This component identifies codeable conditions within clinical notes, surfacing text snippets that may justify codes. \footnote{One challenge is that in US outpatient care (but not inpatient care) notes may mention suspected rather than diagnosed conditions, which should be ignored for coding. Similarly, similar consequential rules may exist in other settings, further underscoring the task's complexity \cite{cms_icd-10-cm_2025}.} A key challenge for this step is that clinical language frequently diverges from standardized index terms, and identifying relevant excerpts is non-trivial.  %

\paragraph{Step 2: \af{index navigator}.} This component locates the authoritative coding references in the alphabetical index by mapping text snippets to valid index terms (e.g., `sepsis'). This involves handling synonyms, variant phrasings, and eponyms to propose preliminary candidate codes. The output is a list of candidate codes (e.g. \textit{A22.7}, \textit{T81.44}, etc. for `sepsis'), each associated with a located term (e.g. `anthrax' and `postprocedural sepsis' for the above examples). Since the alphabetical index is not guaranteed to point to an assignable code, the output of this step is only preliminary.

\paragraph{Step 3: \af{tabular validator}.} This component refines and narrows down the candidate codes by applying formal coding rules. It interacts with the ICD hierarchy and chapter-specific guidelines to resolve ambiguities and anatomical specifications, thereby producing a tentative code set.

\paragraph{Step 4: \af{code reconciler}.} This component finalizes the code assignment, applying instructional notes to resolve mutually exclusive codes and ordering conventions. The output is intended to be the most complete and ordered list of codes that reflects the patient encounter, while adhering to medical coding conventions.

\subsection{Implementation}\label{sec:implementation}

We implement the \textit{CLH} framework as an agentic system (see \autoref{fig:agent_system}). We define agents as modular components that can be implemented in various ways, such as standalone language models and retrieval-augmented generation (RAG). 

In our implementation, each agent is instantiated as a distinct inference step using the same backbone model, but with different instructions (via prompts) tailored to their role (see \autoref{app:prompts}). The current implementation relies on models with `thinking-enabled' mode to enable test time computation \cite{snell_scaling_2024, muennighoff_s1_2025}. 

We use the `reasoning' models for improved performance; theoretically the `reasoning traces' could also provide transparency into the model decision process, but at least for the current `reasoning' models this process is not faithful~\cite{ChenBentonEtAl_2025_Reasoning_Models_Dont_Always_Say_What_They_Think,KambhampatiStechlyEtAl_2025_Stop_Anthropomorphizing_Intermediate_Tokens_as_Reasoning_Thinking_Traces,ShojaeeMirzadehEtAl_2025_Illusion_of_Thinking_Understanding_Strengths_and_Limitations_of_Reasoning_Models_via_Lens_of_Problem_Complexity,zhao2025chainofthoughtreasoningllmsmirage}. Specifically, we experiment with three open-weight models of different sizes: \textit{small} (DeepSeek-R1-0528-Qwen3-8B), \textit{base} (DeepSeek-R1-Distill-LLaMA-70B), and \textit{large} (Qwen3-235B-A22B). Additionally, we evaluate two closed OpenAI models, o3-mini and o4-mini, under a HIPAA-compliant use setup. Further implementation details are provided in \autoref{app:implementation_details}.

The \af{evidence extractor (step 1)} operates solely on clinical notes, extracting verbatim text snippets expected to justify coding decisions. For each snippet, we retrieve the top-10 alphabetical terms by embedding snippets and terms into the same semantic space (see appendix \ref{appx:search-index} for details). Next, the \af{index navigator (step 2)} processes each snippet’s set of terms in parallel, selecting the most appropriate terms, which in turn yield a set of candidate codes. These candidate codes, often grouped by ICD chapter, are then passed in parallel to the \af{tabular validator (step 3)} along with chapter-level guidelines,\footnote{The guidelines span 100+ pages; each chapter averages about three pages. Parallel processing lets us input only the chapters relevant to the candidate codes, typically one or two.} receiving input triplets of \{clinical note, chapter guidelines, candidate codes\} to yield tentative assigned codes. Finally, outputs from parallel processing steps are merged, and the \af{code reconciler (step 4)} receives input triplets of \{clinical note, instructional notes, tentative codes\} to verify and finalize the coding assignments, where instructional notes are retrieved by looking up codes in the ICD hierarchy.

\subsection{Medical code taxonomy}
We focus on ICD-10-CM, as (i) ICD-10 is the most widely adopted coding system globally~\cite{teng_review_2023}, (ii) its introduction made the coding task much more laborious and time-consuming,\footnote{It takes a professional inpatient human coder approximately 30 minutes per case on average~\cite{chen_automatic_2021}.} as the number of codes increased more than sevenfold (rising from ~9,000 to ~70,000), and (iii) it is well-maintained with high-quality open-access resources to guide coding decisions.

\subsection{Data}

\paragraph*{MIMIC} is the most widely used open-access database for research on medical coding. We use the popular MIMIC-III \emph{50} split~\cite{mullenbach_explainable_2018} for fine-tuning.

\paragraph*{MDACE}\cite{cheng_mdace_2023} is a dataset of 4,000 human-verified annotations that link ICD codes to supporting evidence spans within clinical notes. The annotations cover 302 inpatient charts and 52 professional-fee charts from MIMIC-III~\cite{johnson_mimic-iii_2016}, including discharge summaries, physician notes, radiology reports, and other clinical document types representative of real-world coding contexts~\cite{alonso_problems_2020}. Inter-annotator agreement after adjudication is high (Krippendorff’s $\alpha=0.97$ for inpatient; $0.96$ for Profee). We use MDACE for evaluation because it includes human-verified evidence spans ideal for LLM evaluation and directly supports ICD-10-CM, our target taxonomy. To our best knowledge, MDACE is the only public dataset that addresses two well-known issues in MIMIC-based benchmarks: the absence of code-to-text links and broader validity concerns about treating MIMIC codes as a gold standard~\cite{searle_experimental_2020, kim_read_2021}. To our knowledge, MDACE is the first and remains among the very few public resources with token-level evidence for long clinical notes in extreme multi-label coding.

\subsection{Evaluation metrics}\label{sec:metrics}
Like previous studies, we report F1 scores (micro and macro), exact match ratio (EMR), and recall for each agent. We pay particular attention to macro F1, which assigns equal weight to every label. \citet{edin_automated_2023} shows that F1 macro is shaped by the long tail of low-frequency labels, where per-label F1 increases with log frequency until roughly 100 training examples and then plateaus. We call this long-tail subset ``rare codes''; gains on rare codes, therefore, translate directly into higher F1 macro.

\subsection{Baselines}

\paragraph{PLM-ICD.} As discussed in \autoref{sec:related}, the state-of-the-art system for this task is still the PLM-ICD model. %
We follow the implementation details provided by~\citet{edin_unsupervised_2024} to reproduce the model and evaluate it on the test split of MDACE. %

\textbf{LLM out-of-the-box.} We also compare the \textit{CLH} approach to the `naive' LLM-based solution: the DeepSeek R1 distilled Llama3.3-70B that is provided with the set of possible codes in the prompt. Following the PLM-ICD setup, in this experiment, the model is provided with only the subset that occurs in the MDACE dataset and not the full set of ICD-10 codes.  

\section{Results}
\label{sec:results}

\subsection{End-to-end evaluation}
We evaluated our implementation of the \textit{CLH} framework end-to-end, comparing it directly to baselines on MDACE. \autoref{tab:main} summarizes the results. With the largest base model, \textit{CLH} achieves comparable performance with PLM-ICD, while handling a 70-fold larger label space.

Notably, while PLM-ICD achieves higher performance on frequent codes (as indicated by its superior F1 micro score), this advantage primarily stems from its supervised training regime, which utilizes data from the same intensive care unit (ICU) as the test set. This shared setting creates strong distributional priors, allowing PLM-ICD to excel specifically on frequent codes within that clinical practice. Conversely, \textit{CLH} does not benefit from such distributional priors, which explains its superior performance on rare codes.

To our knowledge, no prior work has considered the full label space. Hence, when considering the realistic task of human coders, our results can be regarded as the state-of-the-art. In the future, it may be helpful and fairer to compare medical coding systems in the full and label-constrained settings. 

\begin{table}[t]
\centering
\footnotesize
\begin{threeparttable}
\begin{tabular}{p{1.74cm} p{0.5cm} p{0.5cm} c c c}
\toprule
     &   &      & \multicolumn{2}{c}{F1}  & EMR  \\
Model & \#P & \#C         &  Micro   & Macro  &             \\ \midrule
\multicolumn{6}{c}{\textit{constrained label space (prior work)}} \\
PLM-ICD  &  340M & $1K$   & \textbf{0.48}    & 0.25   & 0.02 \\
PLM-ICD  & 340M & $6K$   & 0.46    & 0.21   & 0.02 \\  \midrule
Llama3-70B$^{\dagger}$ & 70B & $1K$ & 0.28 & 0.18 & 0.01 \\ 
\textit{CLH}-small & 8B &  $1K$ &   0.27    & 0.18   & 0.02      \\ 
\textit{CLH}-base  & 70B & $1K$ &   0.38    & 0.24   & 0.02      \\ 
\textit{CLH}-large  & 235B & $1K$ &   0.43    & \textbf{0.28}   & 0.02      \\ 
\textit{CLH}-o3-mini & -- & $1K$ &   0.37    & 0.24   & 0.02      \\ 
\textit{CLH}-o4-mini & --  & $1K$ &   0.41    & 0.27   & 0.02      \\ \midrule
\multicolumn{6}{c}{\textit{full label space (ours, realistic clinical setting)}} \\
\textit{CLH}-base & 70B  & $70K$ &   0.32    & 0.14   & 0.02      \\ 
\bottomrule
\end{tabular}
\begin{tablenotes}
\item[$\dagger$] The DeepSeek-R1 distilled Llama3-70B model.
\end{tablenotes}
\caption{Performance of the \textit{CLH} framework and baselines on the MDACE dataset. ``\#P'' refers to the number of parameters and ``\#C'' to the number of candidate codes used during inference. For PLM-ICD, this includes all MIMIC codes ($\approx6K$) which overlap with MDACE. Llama3 is prompted with all MDACE codes ($\approx1K$). \textit{CLH} scales to the full ICD coding system ($\approx70K$), but is also evaluated in the MDACE constrained 1K setting for comparison.}\label{tab:main}
\end{threeparttable}
\end{table}

\subsection{Performance of the \textit{CLH} agents}

As discussed in \autoref{sec:architecture}, the rationale for introducing the \textit{CLH} framework is the 4-step process used by medical coders at the NHS. To consider whether this process was indeed beneficial to \textit{CLH}, \autoref{tab:agents} reports the results for each component in the \textit{filtered} setting (i.e., the errors made at the previous step of the pipeline are discarded). 

We observe that the \af{index navigator (step 2)} achieves high recall with low precision, suggesting broad coverage of candidate codes. However, the \af{ tabular validator (step 3)} and the \af{code reconciler (step 4)} show much higher precision while maintaining high recall. 

As with any pipeline approach, \textit{CLH} can propagate early mistakes to later agents. A promising direction for mitigating this in future work is self-refinement~\cite{madaan_self-refine_2023} on the same clinical note. After pass $t$, the code reconciler (step 4) outputs a set of codes with rationales. We could append this output to the note as a lightweight scratchpad, then re-invoke all steps once again. The process then repeats for $t{+}1$.

\section{Model Analysis}

While the end-to-end results provide a summary, they do not capture the nuances of the individual stages of the framework. To better understand the limitations of \textit{CLH}, we present five observations to guide future improvements.

\subsection{Are all code-able cues identified?}

\citet{falis_can_2024} observed that GPT-3.5 works well for coding only when code descriptions are verbatim. That setting is rare in practice, where cues embed abbreviations, medical jargon, and implied context. We analyzed the MDACE expert-annotated evidence spans with the text snippets extracted by the \af{index navigator (step 2)}. Then, we measured recall at the level of ICD-10 classification chapters to identify which cues the agent overlooks, and which the retriever fails to link to alphabetical-index terms.

\begin{figure}[!t]
    \centering
    \includegraphics[width=\linewidth]{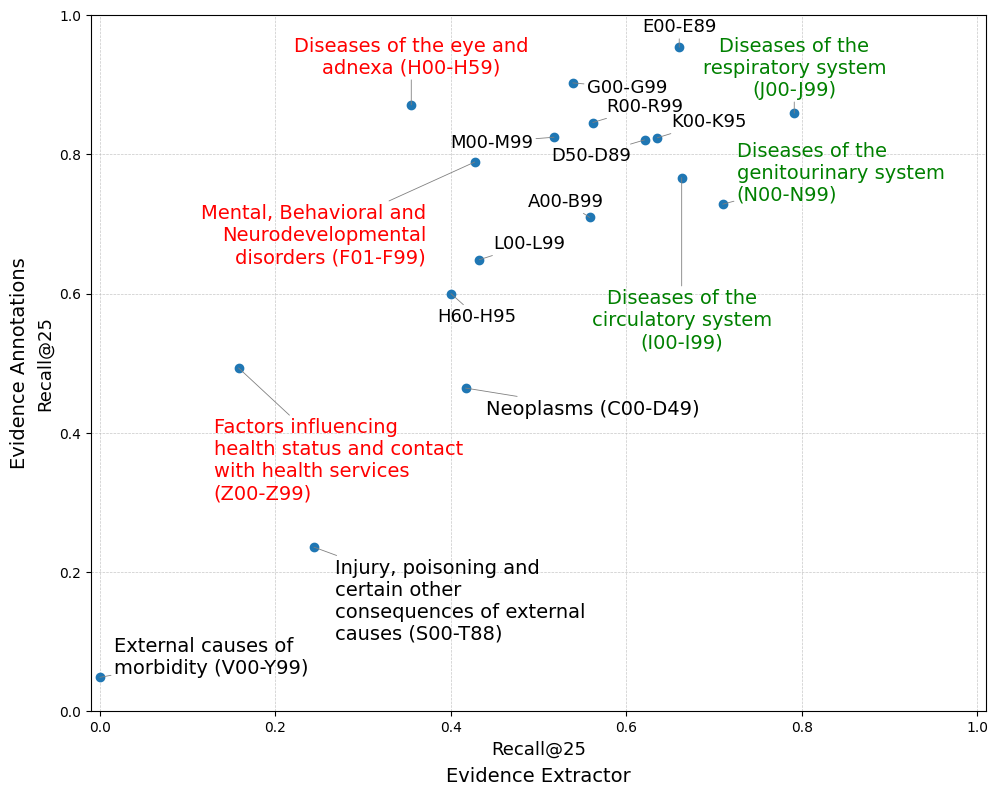}
    \caption{Chapter-level comparison of retrieval recall@$25$. The x-axis reports retrieving alphabetical index terms with the \af{1. evidence extractor}, while the y-axis reports retrieval with expert-annotated evidence.}
    \label{fig:agent_evidence_recall}
\end{figure}

\autoref{fig:agent_evidence_recall} shows the retrieval recall@$25$ for different ICD-10 chapters. X-axis shows recall based on alphabetical terms retrieved using the \af{evidence extractor (step 1)}, while the y-axis reports recall based on expert-annotated text snippets. In each case, we calculate recall for the top 25 retrieved ICD-10 terms. High recall for J00–J99 (respiratory) and N00–N99 (genitourinary) shows that both the agent and expert annotations reliably retrieve index terms for body-system diseases. When the expert annotations recall is limited, as for C00–D49 (neoplasms), S00–T88 (injury or poisoning), and V00–Y99 (external causes), the agent cannot compensate, indicating a retrieval bottleneck. In contrast, we find that for the chapters F01–F99 (mental and behavioral disorders), H00–H59 (eye and adnexa), and Z00–Z99 (factors influencing health and contact with services), the \af{evidence extractor} agent underperforms significantly.

\begin{table}[!t]
\centering
\footnotesize
\resizebox{\linewidth}{!}{%
\begin{tabular}{lcccc}
\toprule
& \multicolumn{2}{c}{F1} & Recall & Precision \\
Model & Micro & Macro &  & \\ \midrule
\af{1. evidence extractor} & 0.12 & 0.09 & 0.62 & 0.06 \\
\af{2. index navigator} & 0.36 & 0.25 & 0.53 & 0.27 \\
\af{3. tabular validator} & 0.40 & 0.27 & 0.47 & 0.34 \\
\af{4. code reconciler} & 0.43 & 0.28 & 0.47 & 0.40 \\
\bottomrule
\end{tabular}%
}
\caption{Performance of the individual components of the \textit{CLH-large} framework on MDACE dataset. Results reflect progressive improvements through the pipeline.}
\label{tab:agents}
\end{table}

To further investigate this shortcoming, we inspected the word clouds for the evidence spans linked to false negatives (see \autoref{fig:z_f_negatives} in the \autoref{appx:retrieval_bound}). 
Frequent misses involve abbreviations (\textit{Hx of CVA}), social or historical qualifiers (\textit{quit smoking}, \textit{tobacco use}), administrative directives (\textit{DNR}), and medication names (\textit{warfarin}, \textit{coumadin}). These cues are not active disease mentions but modifiers that require abbreviation resolution, temporal inference, and context classification. Psychological terms such as \textit{depression} and \textit{anxiety} are likewise overlooked, explaining the agent’s low performance on F01–F99.

Given that much note content is noise~\cite{liu_note_2022}, we speculate that \textit{CLH} could operate on entity+assertion spans rather than full notes, a choice recently shown to maintain accuracy and provide clearer entity-level evidence~\citep{douglas_less_2025}.

\subsection{What is the impact of more candidate codes?}\label{sec:candidate_space_exp}

In this section, we study the effect of the candidate space in a controlled setting that does not rely on \textit{CLH}. At inference time we construct a per-note \emph{candidate set} \(C\) by combining the ground-truth codes \(P=\{ y_i : y_i=1 \}\) with \(K\) hard negatives per positive: \(C=P \cup R_K\) with \(|R_K|=K|P|\). We identify hard negatives by embedding the short descriptions of codes using our retriever (\emph{S-PubMedBert-MS-MARCO}, see \autoref{appx:search-index}) and sampling the \(K\) nearest neighbors that are not in \(P\). This allows us to examine how the size of the candidate set affects model behavior. We then run the \af{tabular validator (step 3)} and the \af{code reconciler (step 4)} only on the candidate set \(C\). The \af{tabular validator (step 3)} receives mutually exclusive choices, which simplifies selection, while the \af{code reconciler (step 4)} operates in a multi-label setting and must also decide how many codes to output. 

\begin{figure}[!t]
    \centering
    \includegraphics[width=0.80\linewidth]{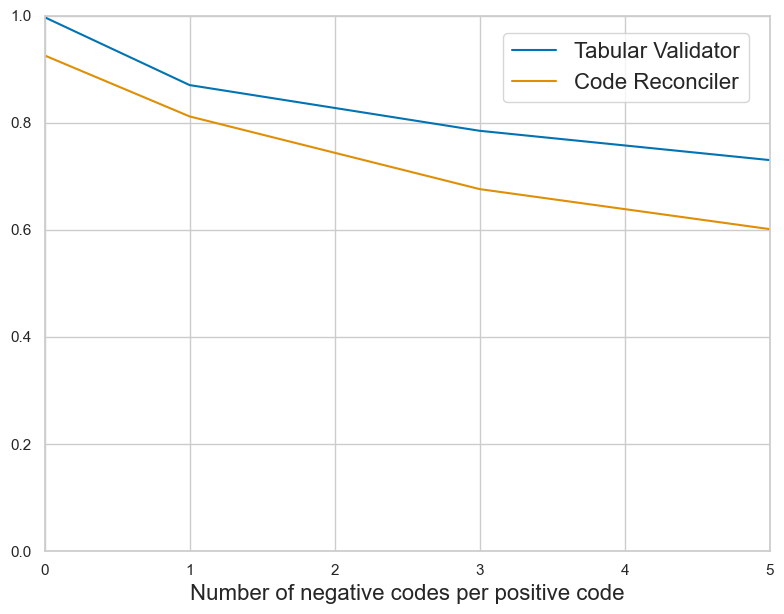}
    \caption{F1 micro scores for the \af{tabular validator (step 3)} and \af{code reconciler (step 4)} as the number of negative codes increases. The \af{code reconciler (step 4)} remains more robust in larger candidate sets due to its simpler prediction task.}
    \label{fig:assign_verify_agent}
\end{figure}

\autoref{fig:assign_verify_agent} shows the F1 micro scores of the \af{tabular validator (step 3)} and \af{code reconciler (step 4)} when scaling the number of negative codes per positive code. As the candidate set grows, both agents exhibit a decline in performance, confirming the challenges of reasoning over larger solution spaces in long, context-rich sequences~\cite{li_long-context_2024}. However, the \af{tabular validator (step 3)} proves more robust, benefiting from a mutually exclusive candidate set that simplifies the selection task. In contrast, the \af{code reconciler (step 4)} must also determine the number of correct codes to output, compounding the difficulty in a multi-label setting. These findings validate our framework's modular design, where the \af{tabular validator (step 3)} predicts tentative codes in parallel, and the \af{code reconciler (step 4)} audits them in a final step.

\subsection{Do the ICD-10 coding resources help?}
Though the context window of some LLMs exceeds millions of tokens \cite{gemini_team_gemini_2024}, they are known to have an \textit{effective} context window of an arbitrary smaller size~\cite{lee_can_2024}. We conducted an ablation study to establish how the capacity of \af{code reconciler (step 4)} saturates with adding more data. We incrementally enriched the input with code-specific information to examine how the model leverages the added context. We started with (1) alphanumeric code identifiers, adding (2) short code descriptions, and then (3) chapter-specific guidelines. We follow the same setup as in \autoref{sec:candidate_space_exp} to construct the candidate set.

\begin{figure}[!t]
    \centering
    \includegraphics[width=0.85\linewidth]{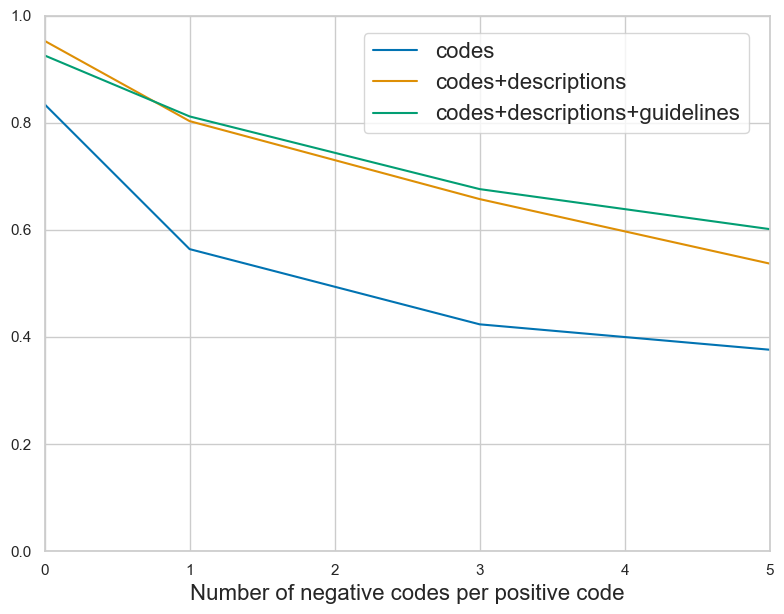}
    \caption{F1 micro scores for \af{3. tabular validator} with added context. Guideline information yields the strongest gains as candidate codes increase.}
    \label{fig:context_window}
\end{figure}

\autoref{fig:context_window} shows that enriching the context input with structured guideline information affects the performance of \af{tabular validator (step 3)}. Starting with only alphanumeric code identifiers, we observe that adding short descriptions leads to clear improvements. Including chapter-specific coding guidelines yields further gains, particularly as the number of candidate codes increases. These results highlight two observations: the model benefits from in-context reasoning grounded in guidelines, and the ability to process a larger candidate set improves with access to richer contextual information about codes. However, as the trend is unlikely to continue infinitely, in practice, the effective size of the context should be identified for a given model.

\subsection{Does `thinking-enabled' decoding help?}

Test-time compute has recently improved accuracy on multiple-choice QA and math benchmarks~\cite{snell_scaling_2024, muennighoff_s1_2025}. We analyzed the performance impact of `thinking-enabled' output generation employing the \af{3. tabular validator}. This setup begins with an unconstrained `reasoning' phase, enclosed within \texttt{<think>...</think>} tags, followed by a constrained output phase within \texttt{<answer>...</answer>} tags. We compared this to structured decoding, where outputs are constrained using a regex pattern to a predefined set of candidate codes~\cite{willard_efficient_2023}, and code prediction begins from the outset of the generation.

\begin{figure}[!t]
    \centering
    \includegraphics[width=0.85\linewidth]{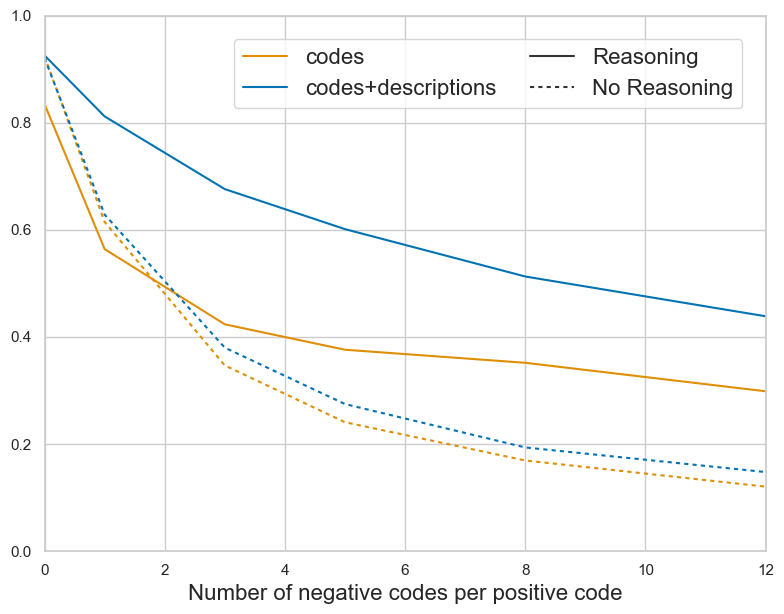}
    \caption{F1 micro scores for the \af{3. tabular validator} with and without reasoning. Reasoning is consistently superior, especially with more negative codes and longer contexts.}
    \label{fig:reasoning_benefit}
\end{figure}

\autoref{fig:reasoning_benefit} compares F1 micro score as the number of negative codes per positive code increases for the \af{3. tabular validator} with and without ‘thinking-enabled’ generation. `Thinking-enabled' generation consistently outperforms structured decoding. Both settings achieve near-perfect accuracy at zero negatives, yet the gap widens as the candidate set and context length grow. The gentler slope suggests the agent can better explore and exploit in-context information throughout the generation process. In contrast, the structured decoding approach initiates code prediction from the outset, limiting the benefit from additional context. It only improves performance marginally when the alphanumeric codes are extended with semantic descriptions.

In the scope of this work, we consider primarily the performance of the `thinking' mode, and do not investigate its interpretability or faithfulness.

\subsection{Can fine-tuning work as well for LLMs as for BERT?}

Despite the growing interest in applying auto-regressive LLMs to clinical text, to our knowledge, no study has fine-tuned them for medical coding. We choose \texttt{Llama-3.2-1B} as our backbone model, and fine-tuned it using LoRA adapters on all the projection matrices in the attention layer and feedforward network. We did not apply any adapters to the language-modeling head. We use supervised fine-tuning, disregarding the order of the codes generated by the model. Due to the limitation of the input size of this model, we can only provide 50 code options in the model prompt. We train and test on MIMIC-III-50, as this allows us to provide descriptions for all 50 codes in the prompt. For efficiency reasons, we subset the training, validation, and test sets to 7500, 2000, and 2000 examples. We train with an effective batch size of 32 for five epochs. We pick the model associated with the best validation F1 macro to produce our test-set metrics (see \autoref{appx:finetune-results} for more details).
\autoref{tab:finetune_compare} shows that fine-tuning a Llama-3.2-1B, a generative model, for medical coding severely underperforms PLM-ICD (a fine-tuned model of BERT generation). 

We speculate that much of PLM-ICD's advantage over predecessor models is due to label-wise attention. We repeat our previous experiment to investigate this empirically, but replace the causal language modeling head with label-wise attention. We follow \citet{edin_unsupervised_2024}, implementing it as cross-attention between all tokens and a learned representation per code. We fully train this in addition to the LoRA updates described above. This changes the task from auto-regressive generation to predicting probabilities over the entire label space simultaneously. We tune the decision boundary based on the highest F1 macro score. We use the boundary associated with the best validation set performance for our test scores. All other fine-tuning details are identical to the auto-regressive setting.

\autoref{tab:finetune_compare} shows that the performance gap is closed when label-wise attention is added to the backbone model, and the performance matches PLM-ICD. It is interesting that even with the label-wise attention, the \texttt{Llama-3.2-1B} model only manages to recover the original performance of MIMIC-III-50 from \citet{edin_automated_2023}, despite being 9 times larger.

\begin{table}[!t]
\centering \footnotesize
\begin{tabular}{lcc}
\hline
 & \multicolumn{2}{c}{\textbf{F1}} \\
\textbf{Model} & \textbf{Micro} & \textbf{Macro}  \\
\hline
PLM-ICD                   & 0.71  & 0.66       \\
Llama-3.2-1B                    & 0.40    & 0.24    \\
+ Label-wise attention            & 0.71   & 0.65     \\
\hline
\end{tabular}
\caption{F1 scores on MIMIC-III-50 for PLM-ICD, fine-tuned Llama-3.2-1B without and with label-wise attention. PLM-ICD scores from \citet{edin_automated_2023}.}
\label{tab:finetune_compare}
\end{table}

\section{Discussion}
\label{sec:discussion}

\subsection{Are LLMs ready for clinical practice?}
\label{sec:present}
Discriminative classifiers, such as PLM-ICD, retain knowledge about every code in their weights, whereas our \textit{CLH} framework dynamically retrieves ICD codes from the alphabetic index, enabling true open-set coding at inference. \textit{CLH} does not yet outperform the PLM-ICD baseline on frequent codes, but it achieves comparable performance, and has several practical advantages. \textit{CLH} is much easier to update when coding conventions change: one would only need to substitute external resources in the prompt. It also suffers less from distributional priors and theoretically can handle even extremely rare codes such as \textit{W59.22 “Struck by a turtle”}~\cite{edin_automated_2023}. 

Nevertheless, both discriminative classifiers and LLMs have apparent limitations that would make both approaches unusable for the real-world ICD-10 end-to-end classification. The former inherit strong distributional priors and only support a toy version of the real ~70K label task. The latter still struggles with precision. Given the current state of the technology, we argue that a more realistic goal for NLP for medical coding is to develop assistive tools that keep human coders in the loop, rather than aim for full end-to-end classification. In that approach, \textit{CLH} has a key advantage, as it already mimics the workflow of human coders.

\subsection{Future work}
\label{sec:future} 
\textbf{Human-computer interaction.} \textit{CLH} framework provides an exciting opportunity to reimagine the task of medical coding as computer-assisted coding, where the goal is not to automate the task but to provide the human coder with the best information to base their decisions on ~\cite{murray_daron_2024}. This could involve research into user interfaces, identification of the pain points where the coders actually want assistance, the threshold of information overload, potential sources of automation bias, and strategies for mitigating them.

\textbf{Improving \textit{CLH} components.} \textit{CLH} allows for more tailored instruction fine-tuning strategies for each component of the pipeline. For example, the \af{evidence extractor (step 1)} can be tuned to predict span-level clinical entities with assertion status and key modifiers (e.g., laterality, acuity)~\cite{douglas_less_2025}, and to pass only asserted, coding-eligible evidence forward, which reduces noise while preserving code-relevant detail. The \af{index navigator (step 2)} can then align entity representations to alphabetical index terms using a label-wise cross-attention objective that improves handling of synonyms and eponyms. The \af{tabular validator (step 3)} could be fine-tuned using clinical notes paired with correct codes and closely related incorrect codes, enabling the model to leverage external resources to identify the most appropriate code. Finally, the \af{code reconciler (step 4)} can be fine-tuned using clinical notes paired with candidate code lists, incorporating hard negatives to train the model in distinguishing relevant from irrelevant codes with greater precision.

\section{Conclusion}
We introduced \textit{Code Like Humans}, a multi-agent framework for medical coding that aligns with human coding practices by integrating previously overlooked resources: the alphabetical index and official guidelines of ICD-10 classification. By relying on these in-context materials rather than only training examples, CLH became the first solution that could support the complete ICD-10 system (over 70K codes). The current implementation outperforms state-of-the-art fine-tuned classifiers on rare codes. A performance gap remains for frequent codes, where fine-tuned classifiers have a natural advantage, but this comes at the expense of addressing only a toy version of the real ICD-10 coding task. 

Our framework offers advantages in easy adaptability to yearly coding updates and opportunities for human-in-the-loop collaboration. Our results suggest that while the current NLP systems can support medical coding, they should aim to complement rather than replace human expertise.

\section*{Limitations}
\label{sec:limitations}

\paragraph{Variety of medical coding practices.} Medical coding varies widely in complexity depending on the coding system (e.g., ICD-9, ICD-10, CPT, etc.), clinical setting (e.g., inpatient, outpatient, professional fee), and document type (e.g., progress notes, radiology reports, discharge summaries). In the scope of this work, we only consider the case of ICD-10 classification, and our evaluation is bounded by the cases considered in the MDACE dataset. More studies considering various models and classification schemes, as well as more public medical benchmark datasets, are needed to address the needs of different healthcare providers. %

\paragraph{How effective is coverage of ICD-10?} Ours is the first work to support the full ICD-10 label space. Still, the current evaluation does not allow for a comprehensive description of how the system would perform in various parts of the taxonomy. The core issue is that there are currently no public evaluation resources that cover the full label space (which is why all prior work has so far focused on a small subset of the codes). This makes it impossible to evaluate any system for performance on all the cases that could be encountered in clinical practice. Development of resources that cover ICD-10 is crucial for further progress in this area.

\paragraph{Supported languages.} Like most other work on clinical NLP, this work focuses on English. There is a growing need for NLP systems to support healthcare assistance in languages other than English, as well as for other medical coding principles and practices worldwide.

\paragraph{Evaluation caveats.} PLM-ICD was fine-tuned on MIMIC-IV ICD-10. Since MDACE contains notes from the same hospital, this could provide an unfair advantage to PLM-ICD over the other models. Conversely, the MIMIC-IV annotations are noisy, where many annotated codes are never mentioned in the clinical notes~\cite{cheng_mdace_2023}. The comparison between PLM-ICD and the other models might have been different if PLM-ICD had been trained on cleaner data from another hospital.

The clinical notes in MDACE and MIMIC have been anonymized (i.e., the special characters have replaced names, dates, addresses, and other identifiers). These special characters may be a source of additional errors for the models.

As elaborated in \autoref{sec:future}, we encourage future work on improving \textit{CLH}, and experimenting with other models is a natural direction to pursue.

\section*{Broader Impacts}
\label{sec:ethics}
\paragraph{Regulation of high-risk AI applications.} According to the enacted EU AI Act~\cite{2024-ai-act}, the list of high-risk applications includes “AI systems intended to be used by public authorities or on behalf of public authorities to evaluate the eligibility of natural persons for essential public assistance benefits and services, including healthcare services, as well as to grant, reduce, revoke, or reclaim such benefits and services”. Where automated medical coding informs eligibility assessment for healthcare services or benefits, it could constitute a high-risk system under the EU AI Act, and carry significant risks to patient care and revenue cycles.

In \autoref{sec:present}, we argued that at the current state of LLM technology, a more realistic and safe approach is to develop methods that assist medical coders rather than provide automated recommendations, while also mitigating the potential automation biases that an assisted workflow may introduce.

\paragraph{Privacy risks.} All work reported in this study was performed using publicly available models and medical datasets, under their existing licenses. We created no new data or base models that could pose new privacy risks.

\paragraph{\textit{CLH} availability.} Our implementation of \textit{CLH} is publicly released under the MIT license, facilitating further research and development. Our solution is intended and marked as a research prototype, not as something that can be deployed out of the box for medical coding applications.

\section*{Acknowledgments}

Generative coding assistance was used in the development of \textit{CLH} implementation.

This work was supported by the Innovation Fund Denmark and Corti ApS through the Industrial PhD program (grant no. 4297-00057B) of Andreas Geert Motzfeldt at the IT University of Copenhagen.

\bibliography{main}

\clearpage
\appendix
\section{External modules}\label{app:external_modules}
\begin{figure}[h!]
    \centering
    \includegraphics[width=0.5\textwidth]{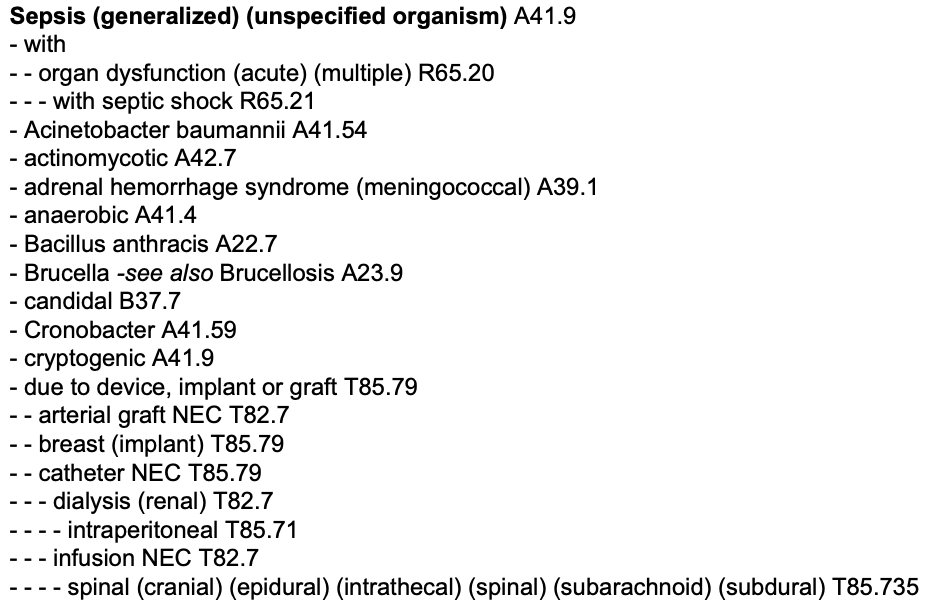}
    \caption{Excerpt from the ICD-10 alphabetical index highlighting the entry term “Sepsis" and indented subterms that guide human coders to the correct term.}
    \label{fig:alphabetical_index_ex}
\end{figure}

\begin{figure}[!ht]
    \centering
    \includegraphics[width=0.5\textwidth]{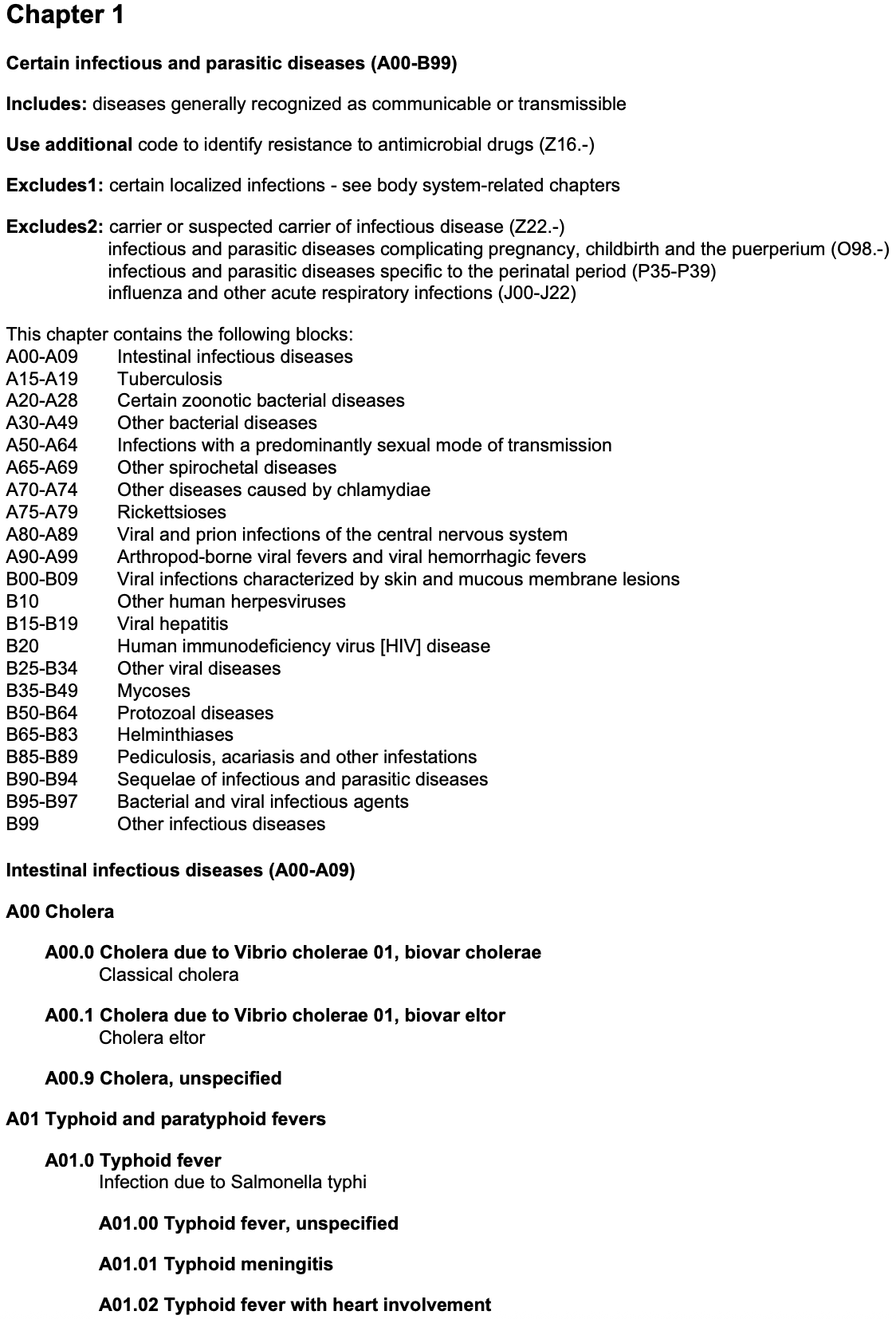}
    \caption{Excerpt from the ICD-10-CM hierarchy (tabular index) showing the beginning of \textit{Chapter 1: Certain infectious and parasitic diseases (A00–B99)}, including instructional notes (includes, use additional, excludes, etc.), followed by hierarchical code blocks and detailed diagnostic categories.}
    \label{fig:tabular_index_ex}
\end{figure}

Here we show the structure of the ICD alphabetic index and hierarchy. The alphabetic index provides an entry point for locating terms and conditions, organized alphabetically and supplemented by subterms and cross-references where each guide term is associated with a code. The ICD hierarchy presents the codes in a hierarchical, chapter-based format organized by body systems or types of conditions. \autoref{fig:alphabetical_index_ex} and \autoref{fig:tabular_index_ex} show representative excerpts from both components to support understanding of the external modules leveraged in our framework.

\subsection{ICD Hierarchy}

\section{Implementation details}\label{app:implementation_details}
Our framework implementation is built on the open-source vLLM inference engine, which enables high-throughput, low-latency serving of LLMs by employing PagedAttention and continuous batching~\cite{kwon_efficient_2023}. vLLM allows us to support long-context inference and concurrent processing efficiently. We power our implementation using four NVIDIA A100 GPUs, each with 80GB of VRAM.

\section{Prompts}\label{app:prompts}
All prompts used in this study were employed with Jinja2-compliant YAML files. This format enables flexible and dynamic prompt construction through template variables. The prompt files are compatible with both completion and chat completion endpoints.
\begin{promptbox1}
You are a highly skilled medical coding assistant trained to extract key lead terms from clinical notes.

Your task is to analyze clinical notes and exhaustively extract the most relevant lead terms and their modifiers.

====== Now let's start! ======
Clinical Note: {{ note }}

Analyze the clinical note and extract the most relevant lead terms. 

Please reason step by step, and output your final answer as a comma-separated list of strings within <answer>...</answer>.
<think>
\end{promptbox1}

\begin{promptbox2}
You are a highly skilled medical coding assistant specializing in structured code extraction from clinical notes.

Your task is to analyze text for medical terminology to output the few most relevant terms from the alphabetical index.

====== Alphabetical Index ======
ID: {{ loop.index }} | Term: "{{ term }}" | ID END: {{ loop.index }}

====== Now let's start! ======
Text: {{ query }}

Analyze the alphabetic index against the text and select the IDs of the most relevant terms. 

If no terms are relevant, output ID 0. 

Please reason step by step, and output your final answer as the IDs of the selected term within <answer>...</answer>.
<think>
\end{promptbox2}

\begin{promptbox3}
You are a highly skilled medical coding assistant specializing in structured code extraction from clinical notes.

Your task is to analyze the clinical note and identify the relevant candidate codes.
    
Note that a clinical note may record many codes, but you are given a subset of candidate codes, where none or one is assignable.

====== Now let's start! ======
Clinical Note: {{ custom_tojson(note | escape) }}

====== Guidelines ======
{{ guidelines }}

====== Candidate Codes ======
ID: {{ loop.index }} |  Code: {{ code }} | ID END: {{ loop.index }}

Analyze the clinical note against the guidelines to select the ID of the most appropriate code. 

If no relevant codes are found, output ID 0. 

Please reason step by step, and output your final answer as the ID of the selected code within <answer>...</answer>.
<think>
\end{promptbox3}

\begin{promptbox4}
You are a highly skilled medical coding assistant specializing in structured code extraction from clinical notes.

Note that a clinical note may record many codes, but you are given a subset of candidate codes where one or more are assignable.

====== Now let's start! ======
Clinical Note: {{ note }}

====== Guidelines ======
{{ guidelines }}

====== Instructional Notes ======
{{ instructional_notes }}

====== Candidate Codes ======
ID: {{ loop.index }} |  Code: {{ code }} | ID END: {{ loop.index }}

Analyze the clinical note against the guidelines and the instructional notes to select the IDs of all assignable codes, prioritizing precision over recall. 

Please reason step by step, and output your final answer as the IDs of the selected codes within <answer>...</answer>.
<think>
\end{promptbox4}

\section{Retrieval results}

\subsection{Dedicated Search Index}\label{appx:search-index}
We employ a vector-based semantic search architecture using Qdrant\footnote{\url{https://qdrant.tech/documentation/}} as the dedicated search index, selected to support sparse, dense, and hybrid search. The alphabetical terms are embedded and stored in a Qdrant index configured with Cosine distance and HNSW indexing (m=32, ef\_construct=256). We use reciprocal rank fusion (RRF) scoring for hybrid search to combine cosine similarity results across models.

\subsection{Retrieval results when using annotated evidence spans.}\label{appx:retrieval_bound}

We simulate an upper bound for our retriever. We use the manually annotated evidence spans from the MDace dataset as search queries. Each span corresponds to codable text snippets identified by human coders and is directly matched against codes in the ICD hierarchy. This setup allows us to evaluate embedding-based models under the assumption of "optimal" codable concept extraction from clinical notes.

\begin{table}[ht]
\setlength{\tabcolsep}{4pt}
\centering
\footnotesize
\resizebox{\linewidth}{!}{%
\begin{tabular}{lcc*{5}{c}}
\textbf{Model} & \textbf{P.} & \textbf{D.} & @5 & @10 & @25 & @50 & @100  \\
\hline
\texttt{bm25} & - & - & 0.52 & 0.56 & 0.63 & 0.67 & 0.70 \\
\texttt{S-PubMedBert} & 110M & 768 & 0.70 & 0.75 & 0.81 & 0.84 & 0.88  \\
\texttt{MedEmbed} & 335M & 768 & 0.62 & 0.67 & 0.76 & 0.81 & 0.86 \\
bm25+\texttt{S-PubMedBert} & 110M & 768 & 0.69 & 0.74 & 0.80 & 0.84 & 0.88\\
\hline
\end{tabular}
}
\caption{Recall@$k$ for different models on lead term search using cosine similarity. P. = Params, D. = Dim.}
\label{tab:expert-annotated-recall}
\end{table}

\begin{figure}[ht]
  \centering
  \includegraphics[width=\linewidth]{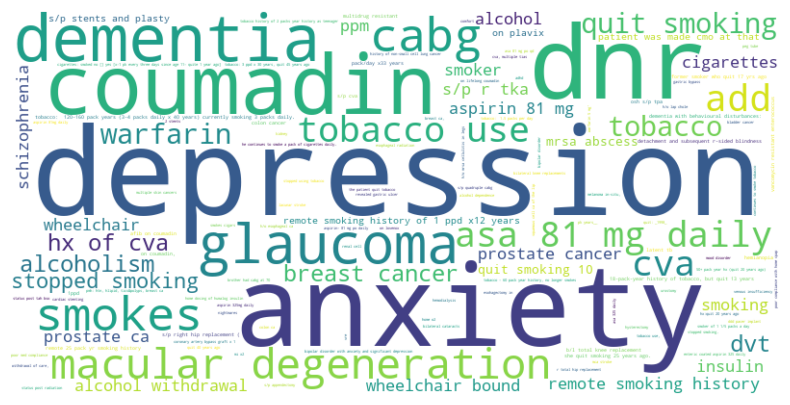}
  \caption{Word-cloud of all expert-annotated spans that the \texttt{Analyze} Agent failed to code in chapters Z00-Z99 (social and administrative factors) and F01-F99 (mental and behavioural disorders). Font size indicates corpus frequency.}
  \label{fig:z_f_negatives}
\end{figure}

\subsection{Retrieval results when using `Analyse' Agent.}

We assess the `Analyse' agent's ability to extract codeable information from clinical notes, using the text snippets it outputs as search queries. The agent processes the clinical note to identify codable information and outputs a list of phrases resembling indexable terms. These phrases serve as input queries to the semantic search index. By comparing recall@$k$ against the results obtained with manually annotated evidence spans, we measure the agent's ability to approximate expert-level identification of codable content. This evaluation captures the compound challenge of accurate information extraction and effective retrieval under real-world constraints.

\begin{table}[ht]
\setlength{\tabcolsep}{4pt}
\centering
\footnotesize
\resizebox{\linewidth}{!}{%
\begin{tabular}{lcc*{5}{c}}
\textbf{Model} & \textbf{P.} & \textbf{D.} & @5 & @10 & @25 & @50 & @100  \\
\hline
\texttt{bm25} & - & - & 0.40 & 0.44 & 0.5 & 0.54 & 0.58 \\
\texttt{S-PubMedBert} & 110M & 768 & 0.50 & 0.54 & 0.60 & 0.64 & 0.68 \\
\texttt{MedEmbed} & 335M & 768 & 0.47 & 0.52 & 0.58 & 0.63 & 0.68 \\
bm25+\texttt{S-PubMedBert} & 110M & 768 & 0.49 & 0.54 & 0.60 & 0.63 & 0.67\\
\hline
\end{tabular}
}
\caption{Recall@$k$ for different models on lead term search using cosine similarity. P. = Params, D. = Dim.}
\label{tab:evidence-extractor-recall}
\end{table}

\subsection{False negative evidence spans for Z00-Z99 and F01-F99}
\label{appx:false_neg_spans}

\autoref{fig:z_f_negatives} visualises the text evidence that the `Analyze' Agent missed when analysing codeable text snippets for chapters Z00–Z99 and F01–F99.

\section{Fine-tuning results}\label{appx:finetune-results}

\subsection{Baseline results}\label{appx:baseline}
We reproduce the PLM-ICD model following the implementation details provided by ~\cite{edin_unsupervised_2024} and train it on MIMIC-IV to evaluate it on the test split of the MDace dataset. This serves as a baseline for comparison with our proposed method. The reproduction adheres to the original model’s architecture and training setup to ensure comparability. PLM-ICD uses a BERT encoder pre-trained on PubMed to encode the text in chunks of 128 tokens, and these contextualized embeddings are fed to a cross-attention layer along with a learned representation for each code.

\begin{table}[h!]
\centering
\caption{Macro and micro F1 scores of the PLM-ICD model by document type on the MDace test split.}
\begin{tabular}{lcc}
\hline
 & \multicolumn{2}{c}{\textbf{F1 Score}} \\
\textbf{Document Type} & \textbf{Macro} & \textbf{Micro} \\
\hline
Consult & 0.09 & 0.20 \\
ECG & 0.29 & 0.56 \\
Nursing & 0.30 & 0.55 \\
Nutrition & 0.17 & 0.26 \\
Case Management & 0.17 & 0.30 \\
Physician & 0.23 & 0.41 \\
Radiology & 0.08 & 0.15 \\
Rehab Services & 0.12 & 0.19 \\
General & 0.21 & 0.35 \\
Discharge Summary & 0.27 & 0.55 \\
\hline
\textbf{Overall} & \textbf{0.21} & \textbf{0.46} \\
\hline
\end{tabular}
\label{tab:f1_scores_by_note_type}
\end{table}

\end{document}